\documentclass[lettersize,journal]{IEEEtran}
\usepackage{amsmath,amsfonts}
\usepackage{algorithmic}
\usepackage{algorithm}
\usepackage{array}
\usepackage[caption=false,font=normalsize,labelfont=sf,textfont=sf]{subfig}
\usepackage{textcomp}
\usepackage{stfloats}
\usepackage{url}
\usepackage{verbatim}
\usepackage{graphicx}
\usepackage{cite}
\usepackage{bm}
\usepackage{multirow}
\usepackage[colorlinks,linkcolor=blue]{hyperref}
\usepackage[capitalize]{cleveref}
\crefname{section}{Sec.}{Secs.}
\Crefname{section}{Section}{Sections}
\Crefname{table}{Table}{Tables}
\crefname{table}{Tab.}{Tabs.}
\def\etal{\emph{et al}.}
\def\ie{\emph{i.e.}}

\usepackage{color}
\usepackage[normalem]{ulem}
\newcommand{{\add}}[1]{\textcolor{blue}{#1}}

\begin{document}

\title{Multi-Scale Invertible Neural Network for Wide-Range Variable-Rate Learned Image Compression}

\author{Hanyue~Tu, Siqi~Wu,
        Li~Li,~\IEEEmembership{Member,~IEEE},
        Wengang~Zhou,~\IEEEmembership{Senior Member,~IEEE},
        and~Houqiang Li,~\IEEEmembership{Fellow,~IEEE}
        % <-this % stops a space
\thanks{This work was supported in part by the Natural Science Foundation of China under Grants 62171429 and 62021001, and in part by the Fundamental Research Funds for the Central Universities under Grant WK3490000007. It was also supported by the GPU cluster built by MCC Lab of Information Science and Technology Institution, USTC.}
\thanks{Hanyue Tu, Li Li, Wengang Zhou and Houqiang Li are
	with the National Engineering Laboratory for Brain-Inspired Intelligence Technology and Application, University of Science and Technology of China, Hefei, 230027, China (e-mail: tuhanyue@mail.ustc.edu.cn; lil1@ustc.edu.cn; zhwg@ustc.edu.cn;
lihq@ustc.edu.cn).}% <-this % stops a space
\thanks{Siqi Wu is
	with the Department of Electrical Engineering, University of Missouri, Columbia, MO 65211 (email: siqiwu@mail.missouri.edu).}% <-this % stops a space
\thanks{Corresponding authors: Li Li and Houqiang Li.}}

\markboth{IEEE TRANSACTIONS ON MULTIMEDIA}%
{Tu \MakeLowercase{\textit{et al.}}: Multi-scale Invertible Neural Network for Wide-range Variable-rate Learned Image Compression}

\IEEEpubid{0000--0000/00\$00.00~\copyright~2021 IEEE}

\maketitle

\begin{abstract}
Autoencoder-based structures have dominated recent learned image compression methods. However, the inherent information loss associated with autoencoders limits their rate-distortion performance at high bit rates and restricts their flexibility of rate adaptation. In this paper, we present a variable-rate image compression model based on invertible transform to overcome these limitations. Specifically, we design a lightweight multi-scale invertible neural network, which bijectively maps the input image into multi-scale latent representations. To improve the compression efficiency, a multi-scale spatial-channel context model with extended gain units is devised to estimate the entropy of the latent representation from high to low levels. 
% Besides, we incorporate extended gain units with the multi-scale structure to enable flexible rate adaptation. 
Experimental results demonstrate that the proposed method achieves state-of-the-art performance compared to existing variable-rate methods, and remains competitive with recent multi-model approaches. Notably, our method is the first learned image compression solution that outperforms VVC across a very wide range of bit rates using a single model, especially at high bit rates. The source code is available at \href{https://github.com/hytu99/MSINN-VRLIC}{https://github.com/hytu99/MSINN-VRLIC}.
\end{abstract}


\section{Introduction}
\label{sec:intro}
\IEEEPARstart{I}{mage} compression is a fundamental task in the field of image processing, aiming at reducing the storage and transmission cost while preserving the visual quality of an image. In the past decades, traditional image/video compression standards have been developed, such as JPEG \cite{wallace1992jpeg}, JPEG2000\cite{taubman2002jpeg2000}, HEVC \cite{sullivan2012overview}, VVC \cite{jvet2020vvc}.
The lossy image compression process can be generally divided into three steps, including transform, quantization, and entropy coding. 
Quantization parameters (QP) are used to control the trade-off between compression efficiency and visual quality. 
A wide range of quantization parameters QP values are supported by the traditional image codecs, which allow the users to determine the desired compression level by adjusting the QP values based on their own needs.

\begin{figure}[t]
  \centering
   \includegraphics[width=0.8\linewidth]{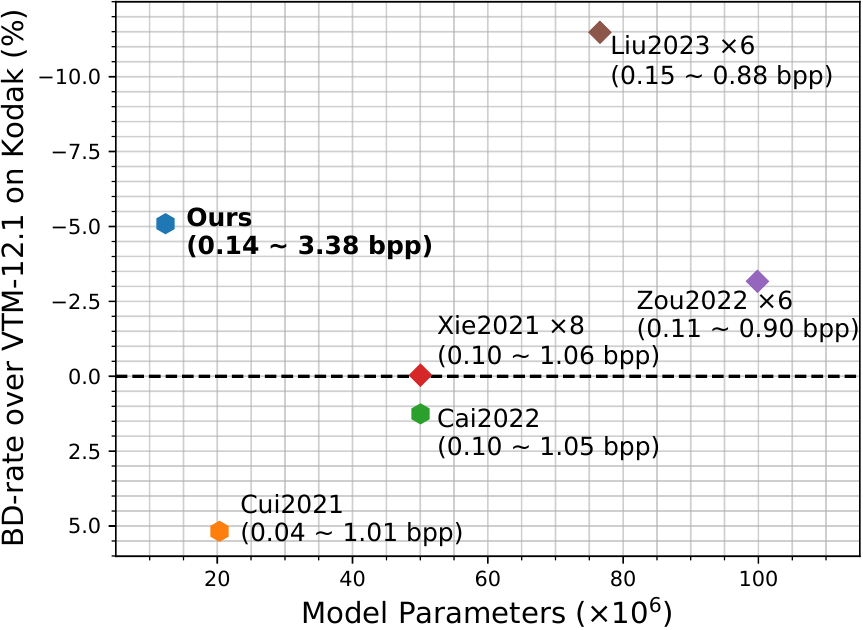}

   \caption{Comparison of compression performance and model size. BD-rate over VTM-12.1 on Kodak is reported. A lower BD-rate indicates better performance.  Our method, Cui2021 \cite{cui2021asymmetric} and Cai2022 \cite{cai2022high} are variable-rate models, while Xie2021 \cite{xie2021enhanced}, Zou2022 \cite{zou2022devil} and  Liu2023 \cite{liu2023learned} are fixed-rate models that require multiple models for different rates (``$\times n$'' means that $n$ models are required). Our method surpasses VTM-12.1 across a wide range of rates with only a single lightweight variable-rate model.}
   \label{fig:fig_1}
\end{figure}

Recently, deep learning-based image compression \cite{balle2017end, balle2018variational, minnen2018joint, cheng2020learned, cui2021asymmetric,  mei2021learning, yang2021towards, 9866778, cai2022high, liu2023learned, gao2023towards, zhang2024learning, llic} has shown promising progress, some of which have outperformed VVC. The state-of-the-art learned image compression methods \cite{zou2022devil, he2022elic, liu2023learned} are usually based on the variational autoencoder \cite{kingma2013auto} architecture. 
% Similar to traditional image codecs, learned-image compression contains neural network-based transform, quantization, and an entropy model, where the transform is composed of an analysis transform (encoder) and a synthesis transform (decoder). In order to improve the compression efficiency, some methods \cite{cheng2020learned, chen2021end, xie2021enhanced, zou2022devil, zhu2022transformer, liu2023learned} utilize more powerful non-linear transform to generate more compressible latent representation, while some other approaches \cite{balle2018variational, minnen2018joint, minnen2020channel, he2021checkerboard, he2022elic} focused on the refining the entropy modeling component. 
Although learned image compression has shown outstanding performance, it still has several limitations. Firstly, most of the state-of-the-art learned image compression methods \cite{he2022elic, liu2023learned} require training multiple fixed-rate models for different target bit rates, which increases the training and storage costs when rate adaptation is needed. Besides, the supported bit rate range of learned image codecs is quite limited when compared with traditional image codecs. To tackle the first issue, variable-rate image compression methods \cite{choi2019variable, cui2021asymmetric, cai2022high} are proposed to utilize a single model for multiple bit rates, where the rate adaptation is achieved by conditioning on the Lagrange multiplier \cite{choi2019variable} or a quality map \cite{song2021variable, cai2022high}, or introducing gain units in latent space \cite{cui2021asymmetric}. Nonetheless, it is worth noting that these variable-rate image compression methods still only support a relatively narrow range of low bit rates.

\IEEEpubidadjcol 
The main reason behind the constraints of autoencoder-based image compression methods is the inevitable information loss caused by the autoencoder itself. The encoder transforms the input image to a more compact latent representation, which has fewer elements than the input image. Therefore, the upper bound of the reconstruction quality is limited inherently. 
An alternative solution is to adopt the invertible neural network \cite{ma2022end, cai2022high} as the transform. With the property of invertibility, the latent representation contains all the information of the input image, and information loss only arises during the quantization process in the latent space. Therefore, the reconstruction quality is not bounded, enhancing the performance at high bitrates. Invertible transform is widely used in traditional image codecs, such as Discrete Cosine Transform (DCT) and Discrete Wavelet Transform (DWT). Previous studies \cite{ma2022end, xie2021enhanced} have explored the development of the end-to-end compression scheme based on invertible transform. However, it is also claimed that invertible neural networks have limited nonlinear transformation capacity.
To address this issue, Ma \etal \cite{ma2022end} relied on a time-consuming RNN-based autoregressive context model, while Xie \etal \cite{xie2021enhanced} introduced extra modules for feature enhancement and dimensionality reduction, which disrupts the bijective property between the input image and the latent representation. Nevertheless, current invertible transform-based learned image compression methods still exhibit inferior performance when compared to autoencoder-based approaches. The potential of adopting invertible transforms in learned image compression remains largely unexplored.
% These methods shown inferior performance compared to autoencoder-based approaches. The application of invertible transform in learned image compression has not been fully explored.
% However,  methods still shown inferior performance compared to autoencoder-based approaches.

% To address this issue, Ma \etal \cite{ma2022end} adopted a time-consuming RNN-based autoregressive context model, while Xie \etal \cite{xie2021enhanced} introduced extra modules for feature enhancement and feature dimensionality reduction, which disrupts the bijective property between the input image and the latent representation. 

In this paper, we present a lightweight variable-rate image compression model based on the invertible neural network, which could cover a wide range of bit rates, ranging from 0.1 to 3.4 bpp on Kodak \cite{kodak1999} with a single model. As shown in \cref{fig:pipeline}, the invertible transform maps the input image to multi-scale latent representations, which are quantized and mapped back to the pixel domain through the reverse of the forward transform.
% The invertible architectures are composed of a series of invertible units. Inspired by the generative method \cite{kingma2018glow}, we utilize an invertible $1\times1$ convolution layer in each unit, which fuses the local channel-wise information and enhances the expressive power of the transformation.
The multi-scale structure reduces the parameters of the network without sacrificing performance and effectively preserves high-frequency information in low-level latent representation.
To establish an efficient and practical compression scheme, we propose a multi-scale spatial-channel context model for accurate entropy modeling. Specifically, multi-layer masked convolutional layers are designed to enlarge the receptive field of the spatial context model while maintaining causality. In addition, to achieve flexible rate adaption, we extend the gain units proposed in \cite{cui2021asymmetric} to better align with our multi-scale structure.
 
We evaluate the proposed compression scheme on Kodak image set \cite{kodak1999}, CLIC professional validation dataset \cite{clic}, and old Tecnick dataset \cite{asuni2014testimages}.  The experimental results demonstrate that our method outperforms VVC across a wide range of bit rates, using a single model, and also shows competitive performance with the state-of-the-art multi-model learned image compression methods. Another advantage of invertible image compression methods is that it is robust for multiple re-encoding, as demonstrated in \cite{helminger2020lossy, cai2022high}.  Our method also outperforms previous compression methods in terms of the fidelity for re-encoding.

The main contributions of our paper are summarized as follows:
\begin{itemize}
    \item 
    Instead of utilizing the commonly-used autoencoder-based architecture, we adopt invertible transform to develop a lightweight learned image codec that supports a wide range of bitrates with high compression efficiency.
    
    % We propose a variable-rate image compression model based on the invertible transform, which is the first learned image compression method that surpasses VVC across a wide range of bit rates using a single model, especially at very high bit rates, \eg, over 3 bits per pixel. 
    \item 
    We design a multi-scale invertible neural network (INN) by stacking a series of simple yet powerful invertible units, which include an invertible convolutional layer and an affine coupling layer to enhance the model's representational capacity.
The multi-scale structure reduces the parameters of our model without sacrificing performance.

% and effectively preserves high-frequency information in low-level latent representation. 
    % We construct a lightweight multi-scale invertible neural network for image compression, which is composed of a sequence of simple invertible units and has only $12M$ parameters.
    
    \item To ensure efficient compression of the multi-scale latent representation, a multi-scale spatial-channel context model with multi-layer masked convolutional layers is designed for accurate entropy modeling.
    
    % a multi-scale spatial-channel context model based on multi-layer masked convolutions is designed for accurate entropy modeling.
    \item 
     The proposed method surpasses VTM-12.1 across a wide range of bit rates using a single model with only $12M$ parameters, especially at very high bit rates. With the property of invertibility, our method also shows the best performance on maintaining rate-distortion performance after multiple re-encodings.
    % With the property of invertibility, our method also shows the best performance on maintaining rate-distortion performance after multiple re-encodings.
\end{itemize}

\section{Related Work}
\label{sec:related_work}

\subsection{Learned Image Compression}

In recent years, there has been remarkable progress in leveraging deep learning for image compression. Most advanced end-to-end image compression methods are built upon the autoencoder architecture \cite{balle2017end, balle2018variational, minnen2018joint, cheng2019energy, balle2020nonlinear}. Ball\'e \etal \cite{balle2017end} proposed the first autoencoder-based image compression framework and utilized additive uniform noise as a continuous approximation of quantization to enable end-to-end training. 
A lot of follow-up methods improve the rate-distortion performance by using more accurate entropy coding modeling and more advanced non-linear transformation. 
The hyperprior \cite{balle2018variational} is introduced to model the entropy model as a conditioned Gaussian distribution. Minnen \etal \cite{minnen2018joint} utilized an autoregressive context model to capture the spatial correlation of latent representations. Li \etal \cite{li2020spatial} used 3D masks to extract spatial-context causal context autoregressively. Channel-wise context model \cite{minnen2020channel} and checkerboard context model \cite{he2021checkerboard, he2022elic, lu2022high} are introduced to improve the computational efficiency of context prediction. Ball\'e \etal \cite{balle2020nonlinear} presented an overview of nonlinear transform coding with both theoretical analysis and empirical study. Wang \etal \cite{wang2021semantic} utilized Laplacian pyramid structure and compressed multi-scale residual images for better texture restoration. Akbari \etal \cite{akbari2021learned} proposed to employ octave-based residual blocks in image compression framework. Recently, attention mechanism \cite{chen2021end, cheng2020learned, zou2022devil, kim2022joint} and transformer-based blocks \cite{lu2022high, qian2022entroformer, zhu2022transformer, koyuncu2022contextformer} are incorporated into the network architectures to further enhance image compression performance. Liu \etal \cite{liu2023learned} proposed a parallel transformer-CNN mixture block and incorporated a transformer-based attention module into the channel-wise entropy model. 

% Some other studies also attempted to improve the compression efficiency from the perspective of vector quantization \cite{zhu2022unified, zhang2023lvqac} and data-dependent transform \cite{wang2022neural}. 

Most learned image compression methods need to train and store multiple models for different rate operating points. Several studies have investigated the potential of using one single model to achieve \emph{variable-rate} compression. Theis \etal \cite{theis2017lossy} introduced and finetuned scale parameters for different rates while keeping the parameters of the pre-trained autoencoder fixed. Choi \etal \cite{choi2019variable} proposed a conditional autoencoder, which is conditioned on the Lagrange multiplier. Besides, mixed quantization training is utilized in \cite{choi2019variable} to allow finer rate adaptation by tuning the quantization bin size. Song \etal \cite{song2021variable} proposed a quality map-guided compression model to support spatially-adaptive variable rates. Sun \etal \cite{sun2021interpolation} introduced an interpolation channel attention module to achieve fine rate control. Cui \etal \cite{cui2021asymmetric} designed gain units to rescale the latent representations and used the exponent interpolation to enable continuous rate adaptation. Cai \etal \cite{cai2022high} proposed an invertible activation transformation module, which takes the quality level as input to generate element-wise scaling and bias tensors.

\begin{figure*}[t]
  \centering
   \includegraphics[width=0.95\linewidth]{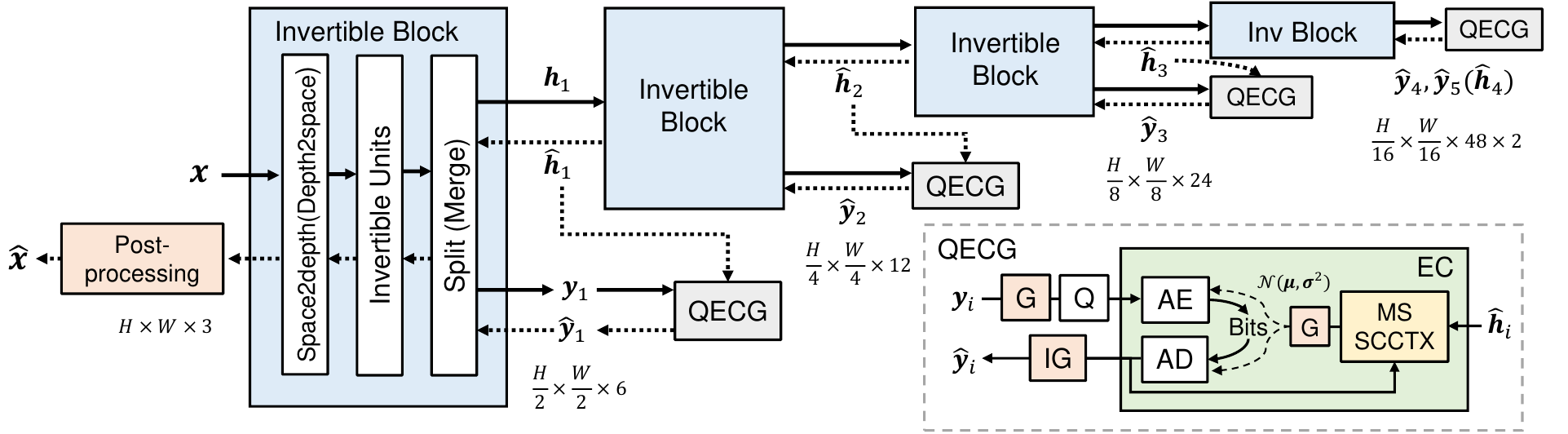}
   \caption{Framework of our proposed invertible transform-based image compression model. Four-level invertible blocks map the input image to multi-scale latent representations $\{\bm{y}_1, \bm{y}_2,\bm{y}_3,\bm{y}_4,\bm{y}_5\}$, which are quantized and passed through the reverse process of the network to reconstruct the image. The post-processing module is used to compensate for the quantization loss. ``QECG'' includes the process of quantization and entropy coding with gain units. ``G'' and ``IG'' denote gain units and inverse gain units, respectively. ``MS-SCCTX'' is the proposed multi-scale spatial-context model.  ``Q'' represents quantization. ``AE'' and ``AD'' stand for arithmetic encoding and decoding, respectively.}
   \label{fig:pipeline}
\end{figure*}

\subsection{Invertible Neural Network}

An invertible neural network is a special kind of neural network, which is composed of several bijective mappings. Therefore, the input can be losslessly recovered given the output through a reverse process of the neural network. Invertible neural networks have been widely studied in the context of normalizing flows \cite{rezende2015variational, dinh2014nice}, which employs a sequence of invertible mapping to transform the complex distributions of input data into simpler ones. The commonly used invertible architectures are additive coupling layers \cite{dinh2014nice} and affine coupling layers \cite{dinh2016density}. Instead of directly applying the transformation to the input $\mathbf{x}$, the input $\mathbf{x}\in \mathbb{R}^D$ is first split into $(\mathbf{x}_1, \mathbf{x}_2)\in \mathbb{R}^d \times \mathbb{R}^{D-d}$. Then the addictive coupling layer can be formulated as $(\mathbf{y}_1, \mathbf{y}_2) = (\mathbf{x}_1, \mathbf{x}_2 + f(\mathbf{x}_1))$, and the affine coupling layer can be formulated as $(\mathbf{y}_1, \mathbf{y}_2) = (\mathbf{x}_1, \mathbf{x}_2 \odot \exp(s(\mathbf{x}_1)) + f(\mathbf{x}_1))$. The coupling layers are composed in an alternating pattern.  Kingma \etal \cite{kingma2018glow} proposed to replace the fixed permutation in the flow model with a learned invertible 1 × 1 convolution.

Invertible neural networks are demonstrated effective in other research domains as well.  Gomez \etal \cite{gomez2017reversible} proposed reversible residual network (RevNet), where the activations of each layer can be exactly reconstructed from the activations of the next layer. Therefore, the activations for most layers need not be stored in memory. Jacobsen \etal \cite{jacobsen2018revnet} built a fully invertible architecture i-RevNet upon RevNet and showed that it can achieve the same classification accuracy compared to similar non-invertible architectures. Xiao \etal \cite{xiao2020invertible} designed an invertible rescaling network to transform high-resolution images into low-resolution images and a case-agnostic latent variable.

Invertible transform is also widely used in traditional hybrid image codecs, such as Discrete Cosine Transform (DCT), Discrete Wavelet Transform (DWT), and the lifting scheme \cite{sweldens1998lifting}. Recently, some works also developed learned image compression based on invertible architecture. Helminger \etal \cite{helminger2020lossy} explored the potential of normalizing flows for lossy image compression. The model covers a wide range of quality levels but the performance degrades at low bit rates. Ma \etal \cite{ma2020iwave, ma2022end} designed a trainable wave-like transform 
to support both lossless and lossy image compression. Xie \etal \cite{xie2021enhanced}
presented an enhanced invertible encoding network to mitigate the information loss problem, which used an attentive channel squeeze layer and a feature
enhancement module for better compression performance. Li \etal \cite{li2021reversible} designed a reversible autoencoder with a guarantee of
stability for image reconstruction. Cai \etal \cite{cai2022high} introduced invertible activation transformation and extended  \cite{xie2021enhanced} to a variable-rate model.

\begin{figure}[t]
  \centering
   \includegraphics[width=0.8\linewidth]{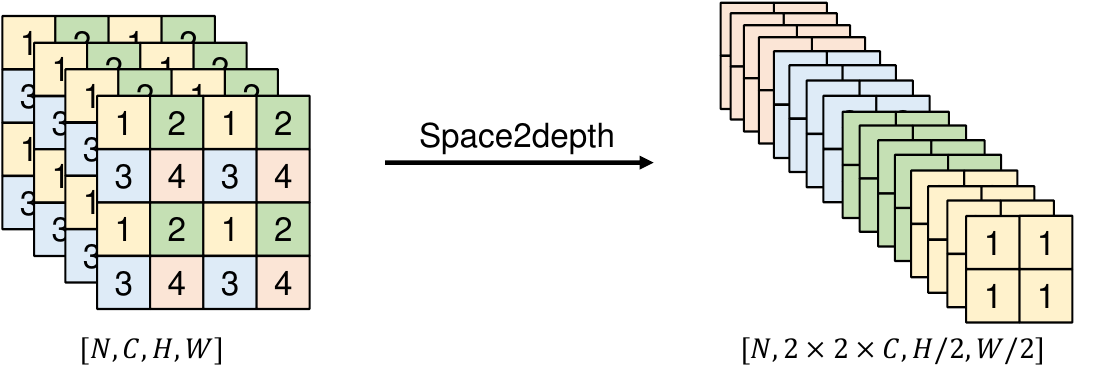}
   \caption{Invertible down-scaling using the space-to-depth module.}
   \label{fig:space2depth}
\end{figure}

\section{Proposed Method}
\label{sec:method}

\subsection{Framework}

The framework of our proposed image compression model based on invertible transform is illustrated in \cref{fig:pipeline}. We employ a multi-scale architecture. The input image $\bm{x}\in \mathbb{R}^{H\times W\times3}$ is first invertibly down-scaled by the space-to-depth \cite{he2021checkerboard} module. We divide the input image into non-overlapping $2\times2$ patches. Each $2\times2$ patch is flattened to a one-dimensional vector through a series of reshaping and permuting operations, as shown in \cref{fig:space2depth}. The down-scaled image $\bm{x}_{ds} \in\mathbb{R}^{\frac{H}{2}\times \frac{W}{2}\times12}$ are then passed through a sequence of invertible units. This sequence enables invertible computations and captures the dependencies within the data. The output is split along the channel dimension, generating the first latent representation $\bm{y}_1$ and another hidden representation $\bm{h}_1$, which is formulated as:
\begin{equation}
    (\bm{y}_1, \bm{h}_1) = IB_1(\bm{x}),
\end{equation}
where $IB$ represents the invertible block. The hidden representation $\bm{h}_1$ is then passed as input to the subsequent invertible block at the next level. This recursive process is defined as follows:
\begin{equation}
\begin{aligned}
     (\bm{y}_{i+1}, \bm{h}_{i+1}) =& IB_{i+1}(\bm{h}_i), 1\leq i\leq2, \\
    (\bm{y}_{4}, \bm{y}_{5}) =& IB_{4}(\bm{h}_3).   
\end{aligned}
\end{equation}
Four-level invertible blocks are used in our model and we obtain the multi-scale latent representations $\mathbf{Y}=\{\bm{y}_1, \bm{y}_2,\bm{y}_3,\bm{y}_4,\bm{y}_5\}$.

The latent representations $\mathbf{Y}$ are quantized and encoded into bitstream. During the image reconstruction phase, the decoded latent representations $\hat{\mathbf{Y}}$ go through the reverse process of the invertible blocks. At the end of the decoding process, a post-processing module is utilized to compensate for the information loss resulting from the quantization. The decoding process is formulated as:
\begin{equation}
    \hat{\bm{x}} = P(IB^{-1}_1(IB^{-1}_2(IB^{-1}_3(IB^{-1}_4(\hat{\bm{y}}_5,\hat{\bm{y}}_4), \hat{\bm{y}}_3),\hat{\bm{y}}_2), \hat{\bm{y}}_1)),
\end{equation}
where $IB^{-1}$ represents the reverse process of the invertible block, and $P$ denotes the post-processing module.  

\begin{figure}[t]
  \centering
   \includegraphics[width=0.95\linewidth]{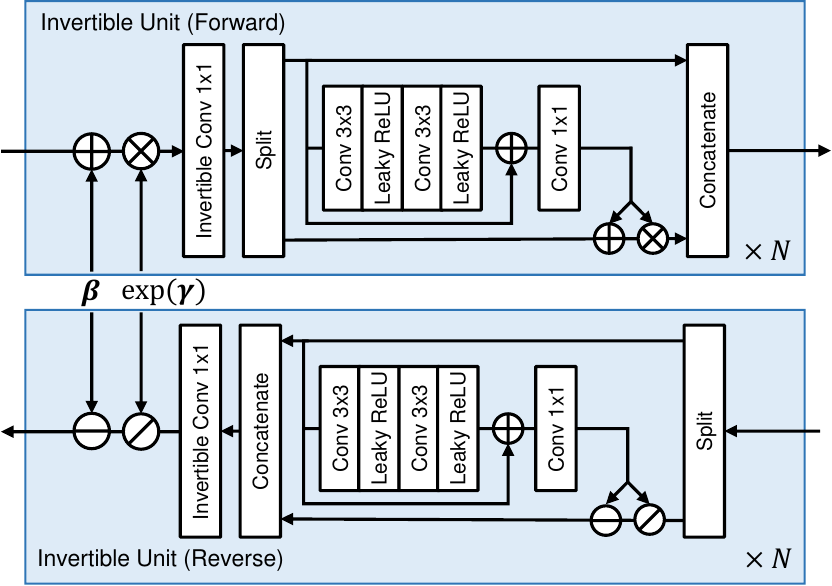}

   \caption{Architecture of the invertible unit. We stack $N$ invertible units within each invertible block.}
   \label{fig:inv_units}
\end{figure}

\subsection{Architecture of Invertible Units}

At each level, the invertible block is composed of $N$ invertible units. The architecture of a single invertible unit is shown in \cref{fig:inv_units}, which is designed based on the flow step in the generative model \cite{kingma2018glow}. In the forward process, given the input tensor $\bm{t}$, a data-independent channel-wise normalization is applied first. Learnable vectors $\bm{\beta}$ and $\bm{\gamma}$ are initialized as $-\text{E}(\bm{t})$ and $-\log(\sqrt{\text{Var}(\bm{t})})$. The mean and variance are calculated over the spatial dimension per channel. 
% We found that the learnable normalization layer can stabilize and accelerate the training process.

Different from previous invertible transform-based image compression methods \cite{ma2022end, xie2021enhanced}, which split the tensor into two chunks and directly input the two chunks into multiple sequential coupling layers, we utilize an invertible $1\times1$ convolutional layer before each coupling layer. The invertible $1\times1$ convolutional layer fuses the local features in the channel dimension, facilitating more frequent exchange of information and enhancing the representation power of the transformation.  After the invertible $1\times1$ convolutional layer, an affine coupling layer is applied. Denoting the input of the coupling layer as $\bm{u}\in \mathbb{R}^{h\times w\times C}$, we first evenly split it into two parts, $\bm{u}_1$ and $\bm{u}_2$, each with a size of ${h\times w\times \frac{C}{2}}$. Then the first tensor $\bm{u}_1$ is fed into a residual block and a $1\times1$ convolutional layer is used to generate the bias and scale parameters that are applied to $\bm{u}_2$. At last, $\bm{u}_1$ and the updated $\bm{u}'_2$ are concatenated again and passed through the next invertible unit. This process can be formulated as follows:
\begin{equation}
\begin{aligned}
\bm{t}' & = (\bm{t} + \bm{\beta}) \otimes \exp(\bm{\gamma}), \\
     \bm{u} & = \bm{t}' \cdot \bm{W}, \\
     \bm{u}_1, \bm{u}_2 & = \text{Split}( \bm{u}), \\
     \text{bias}, \text{scale} & = \text{Split}(\text{Conv}_{1\times1}(\text{ResBlock}( \bm{u}_1))), \\
     \bm{u}'_2 &= (\bm{u}_2 + \text{bias}) \odot \exp(2\sigma(\text{scale})-1),\\
     \bm{u}' &= \text{Concatenate}(\bm{u}_1, \bm{u}'_2 )
\end{aligned}
\label{eqn:inv1}
\end{equation}
where $\otimes$ denotes channel-wise multiplication, $\odot$ denotes Hadamard product, and $\sigma(\cdot)$ represents the sigmoid function.

Correspondingly, we can formulate the reverse process of the invertible unit as:
\begin{equation}
\begin{aligned}
    (\bm{u}_ 1, \bm{u}'_2 ) &= \text{Split}(\bm{u}'), \\
     \text{bias}, \text{scale} & = \text{Split}(\text{Conv}_{1\times1}(\text{ResBlock}( \bm{u}_1))), \\
     \bm{u}_2 &= \bm{u}'_2 \odot \exp(1-2\sigma(\text{scale})) -\text{bias},\\
     \bm{u} &= \text{Concatenate}(\bm{u}_1, \bm{u}_2 ) \\
     \bm{t}' & = \bm{u} \cdot \bm{W}^{-1}, \\
     \bm{t} &= \bm{t}' \otimes \exp(-\bm{\gamma}) - \bm{\beta},
\end{aligned}
\label{eqn:inv2}
\end{equation}

\begin{figure}[t]
  \centering
   \includegraphics[width=0.95\linewidth]{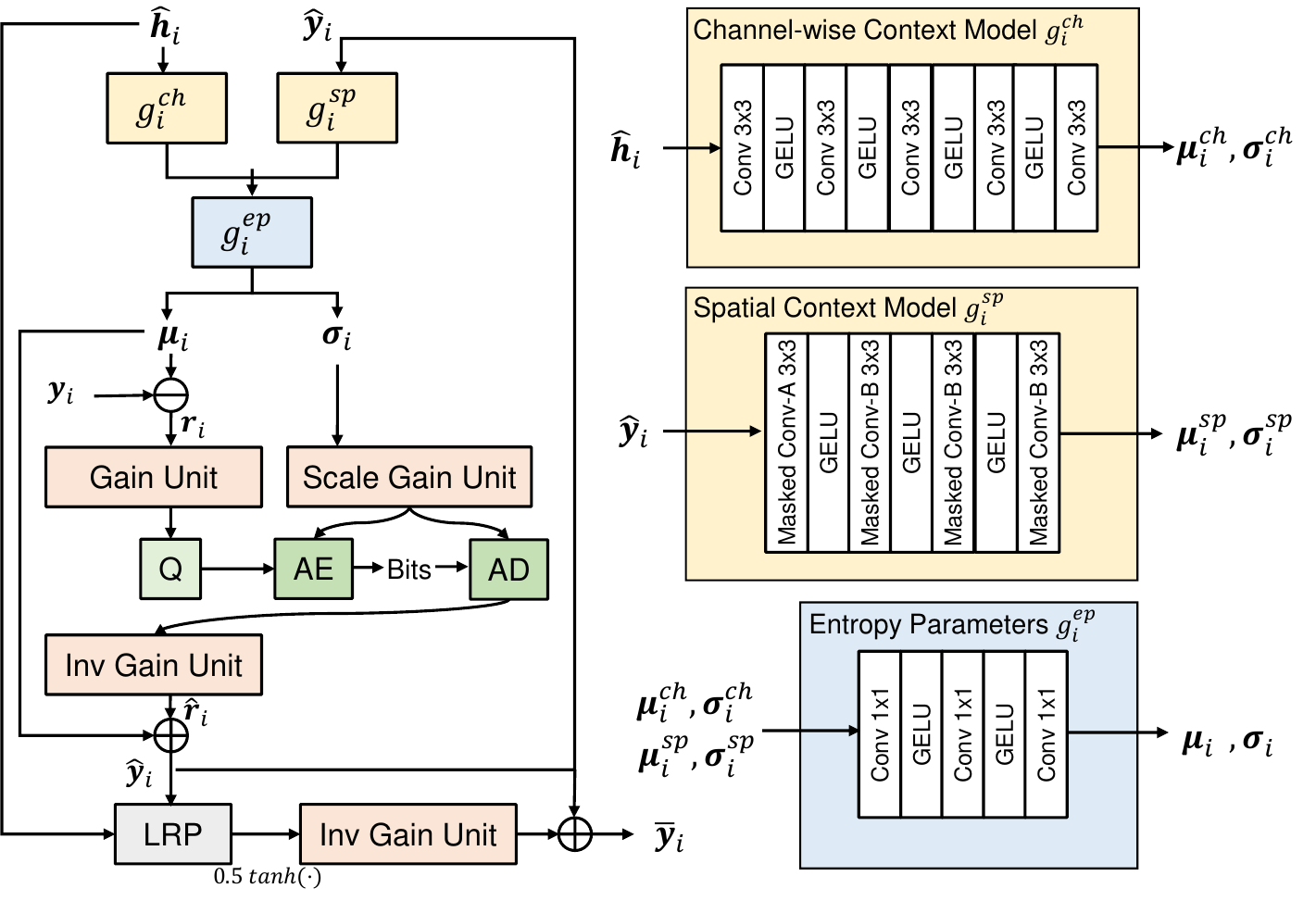}
   \caption{Proposed spatial-channel context model with extended gain units for entropy coding. The architecture of the latent residual prediction (LRP) module is the same as the channel-wise context model.}
   \label{fig:entropy}
\end{figure}

% In our implementation, we set the number of invertible units in each blocks $N=4 $.

\subsection{Multi-scale Spatial-channel Context Model}
\label{sec:context_model}

Previous invertible transform-based image compression methods \cite{ma2022end, xie2021enhanced, cai2022high} adopted the time-consuming serial auto-regressive context models,  which are actually impractical for establishing an effective compression scheme. In this paper, we design a multi-scale spatial-channel context model to efficiently encode the multi-scale latent representations $\mathbf{Y}=\{\bm{y}_1, \bm{y}_2,\bm{y}_3,\bm{y}_4,\bm{y}_5\}$. 
% Note that we do not employ the commonly used hyperprior architecture \cite{balle2018variational}, which involves a hyper encoder to generate the hyper latent representation from $\mathbf{Y}$ as side information and a hyper decoder to generate the entropy parameters for compressing $\mathbf{Y}$. 
% \subsubsection{Multi-scale channel-wise context model}
Our method encodes the latent representations $\mathbf{Y}$ from low resolution to high resolution. The entropy model for quantized latent $\hat{\bm{y}}_i$  is illustrated in \cref{fig:entropy}. Each element in $\hat{\bm{y}}_i$ is modeled as a single Gaussian distribution. 
% We assume that $\hat{\bm{y}}_5$ follows a zero-mean Gaussian distribution, whose scale is defined as learnable parameters. 
% The distribution is shared among all elements with the same channel. 
% Then we continue to encode the latent codes $\hat{\bm{y}}_i (i \le 4)$. 
The channel-wise Gaussian parameters mean $\bm{\mu}_i^{ch}$ and scale $\bm{\sigma}_i^{ch}$ are conditioned on the previously decoded lower-resolution latent codes $\{\hat{\bm{y}}_{>i}\}$.  For slice $\hat{\bm{y}}_5$, $\bm{\mu}_5^{ch}$ is set to zero and $\bm{\sigma}_5^{ch}$ is defined as learnable parameters. To align the resolution for channel-wise context prediction, we utilize the reverse process of the invertible transform to reconstruct the hidden representation $\hat{\bm{h}}_i$, which has the same shape as $\hat{\bm{y}}_i$. The channel-wise context model $g_i^{ch}$ takes $\hat{\bm{h}}_i$ as input and produces the channel-conditional $\bm{\mu}_i^{ ch}$ and $\bm{\sigma}_i^{ch}$.

\begin{figure}[t]
  \centering
   % \includegraphics[width=0.8\linewidth]{fig/mask_conv.pdf}
   \subfloat[]{
    \includegraphics[width=0.25\linewidth]{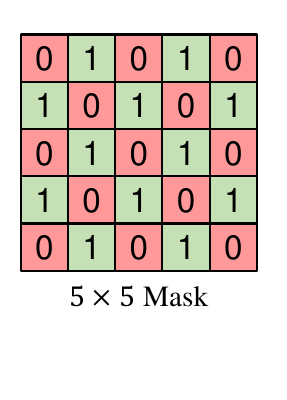}
    \label{fig:mask1}}
     \subfloat[]{
    \includegraphics[width=0.21\linewidth]{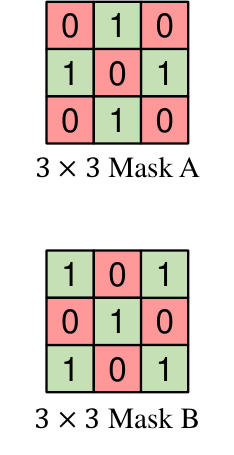}
    \label{fig:mask2}}
  \subfloat[]{
    \includegraphics[width=0.48\linewidth]{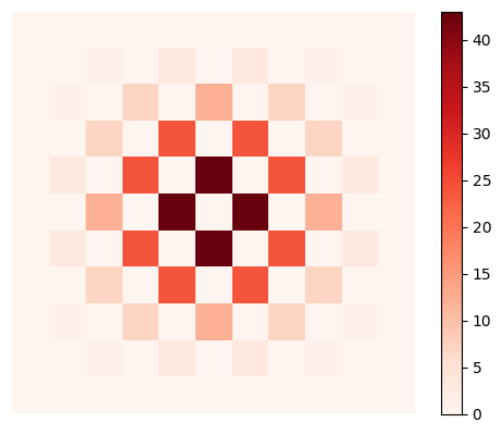}
    \label{fig:mask3}}
   \caption{(a) $5\times5$ mask used in \cite{he2021checkerboard, he2022elic}. (b) Two types of $3\times3$ masks used in our spatial context model. (c) Visualization of the weighted receptive field of our spatial context model, which consists of one masked convolution-A and three masked convolution-B.}
   \label{fig:mask}
\end{figure}

To reduce the spatial redundancy, we further incorporate the checkerboard context model proposed in \cite{he2021checkerboard} into the entropy model of each latent code. The latent code $\hat{\bm{y}}_i$ is split into two parts, $\hat{\bm{y}}_i^{\text{anchor}}$ and $\hat{\bm{y}} _i^{\text{non-anchor}}$, in a checkerboard pattern, and the entropy model for $\hat{\bm{y}}_i^{\text{non-anchor}}$ is conditioned on $\hat{\bm{y}}_i^{\text{anchor}}$. Previous methods \cite{he2021checkerboard, he2022elic} utilized a single-layer $5\times5$ checkerboard-masked convolution to model the spatial correlation, as shown in \cref{fig:mask1}. This is sufficient in previous autoencoder-based models with lower-resolution latent representations, where each element corresponds to 16 pixels in the input image. However, in our multi-scale architecture,  the low-level latent representation is larger, requiring a context model with a larger receptive field.

We design a multi-layer spatial context model by stacking four $3\times3$ masked convolutional layers. Specifically, the first layer uses mask A while the consecutive layers use mask B, as shown in \cref{fig:mask2}. We visualize the weighted receptive field of our multi-layer masked convolutional layers in \cref{fig:mask3}. To visualize the receptive field, we set all the 
convolution weights to 1 and biases to 0, and then calculate the gradient magnitude of the input with respect to the output center pixel. Our multi-layer spatial context model preserves the causality while benefiting from a larger receptive field and increased nonlinearity. Finally, the entropy parameter model  $g_i^{ep}$ aggregates the estimated parameters from $g_i^{ch}$ and $g_i^{sp}$. In summary, our context model can be formulated as:
\begin{equation}
\begin{aligned}
\bm{\mu}_5, \bm{\sigma}_5 &=  g_5^{ep}(\bm{0}, \bm{\sigma}_5^{ch}, g_5^{sp}(\hat{\bm{y}}_i^{\text{anchor}})), \\
\bm{\mu}_i, \bm{\sigma}_i &=  g_i^{ep}(g_i^{ch}(\hat{\bm{h}}_i), g_i^{sp}(\hat{\bm{y}}_i^{\text{anchor}})), 1\le i\le4, \\
\hat{\bm{h}}_i &= IB^{-1}_{i+1}(\hat{\bm{h}}_{i+1},\hat{\bm{y}}_{i+1})=IB^{-1}_{>i}(\hat{\bm{y}}_{>i})),
\end{aligned}
\end{equation}
where $IB^{-1}_{>i}$ represents the reverse process of the transform starting from the $i+1$-th invertible blocks onwards. The spatial context term $g_i^{sp}$ is set to zero for all the elements in $\hat{\bm{y}}_i^{\text{anchor}}$.

\subsection{Rate Adaptation with Extended Gain Units}

Gain units are introduced in \cite{cui2021asymmetric} to achieve continuous rate adaptation by rescaling the magnitude of the latent representation using channel-wise multiplication before quantization and entropy parameter estimation. However, the original gain units are unsuitable for our multi-scale architecture because our channel-wise context model takes as input the reconstructed hidden representation $\hat{\bm{h}}_i$, instead of the scaled latent representations, as explained in \cref{sec:context_model}. The inconsistency in magnitude between the input and output can adversely affect the training process of the channel-wise context model. In our approach, the entropy parameters are all calculated based on the unscaled hidden and latent representations. We place the gain units and the inverse gain units immediately before and after the quantization process, which is formulated as:
\begin{equation}
        \hat{\bm{y}}_{i,q} =  IG_i(Q(G_i(\bm{y}_i - \bm{\mu}_i, q)), q) + \bm{\mu}_i
\end{equation}
where $Q(\cdot)$ represents the quantization process, $G_i(\cdot)$ and  $IG_i(\cdot)$ represent the gain process and inverse gain process, respectively, and $q$ denotes the quality value, which can take any arbitrary value within a predefined range using the exponential interpolation \cite{cui2021asymmetric}.

In practical implementation, it is the gained residual $Q(G_i(\bm{y}_i - \bm{\mu}_i, q))$ to be encoded, instead of $\hat{\bm{y}}_{i,q}$. The quantized gained residual is modeled as a single Gaussian distribution. In order to estimate the rates in a variable-rate model, we also apply the gain process to the estimated scale parameter $\bm{\sigma}_i$, as below:
\begin{equation}
\begin{aligned}
    P(\hat{\bm{y}}_{i,q}|\hat{\bm{y}}_{>i,q})&\sim P(Q(G_i(\bm{y}_i - \bm{\mu}_i, q))|\hat{\bm{y}}_{>i,q}) \\ &\sim \mathcal{N}(\bm{0}, G'_i(\bm{\sigma}_i, q))
\end{aligned}
\end{equation}
Latent residual prediction (LRP) \cite{minnen2020channel} aims to reduce the quantization error by predicting the residual in latent space. The predicted residual is scaled by the inverse gain units because the quantization error of $\hat{\bm{y}}_i$ falls in the range of $(IG_i(-0.5, q), IG_i(0.5, q))$.
% \begin{equation}
%     P(Q(G_i(\bm{y}_i - \bm{\mu}_i)) &\sim \mathcal{N}(\bm{0}, G'_i(\bm{\sigma}_i))
% \end{equation}

During the training process of our model, we randomly sample the quality values $q$ from the predefined set $\mathcal{S}_q=\{0, 1, 2, ..., q_{max}\}$. Each quality value $q$ corresponds to a Lagrange multiplier $\lambda_q$ in the rate-distortion optimization problem.  The loss function of the proposed variable-rate model is defined as:
\begin{equation}
\label{eqn:loss}
\begin{aligned}
    \mathcal{L} = & \sum_{q\in \mathcal{S}_q} \sum_{i=1}^{5}R(\hat{\bm{y}}_{i,q}) + \lambda_q D(\bm{x}, \hat{\bm{x}}_q) \\
    = & \sum_{q\in \mathcal{S}_q} \sum_{i=1}^{4} \mathbb{E}[-\log_2(P(\hat{\bm{y}}_{i,q}|\hat{\bm{y}}_{>i,q})] \\
    & + \mathbb{E}[-\log_2(P(\hat{\bm{y}}_{5,q})] + \lambda_q D(\bm{x}, \hat{\bm{x}}_q),
\end{aligned}
\end{equation}
where $D(\cdot)$ represents the distortion term such as mean squared error (MSE). During the inference time,  we can select a specific quality value $q\in[1, q_{max}]$ to achieve the desired compression bit rate.

  \begin{figure*}[t]
  \centering
   % \includegraphics[width=0.8\linewidth]{fig/mask_conv.pdf}
   \subfloat[Kodak]{
    \includegraphics[width=0.31\linewidth]{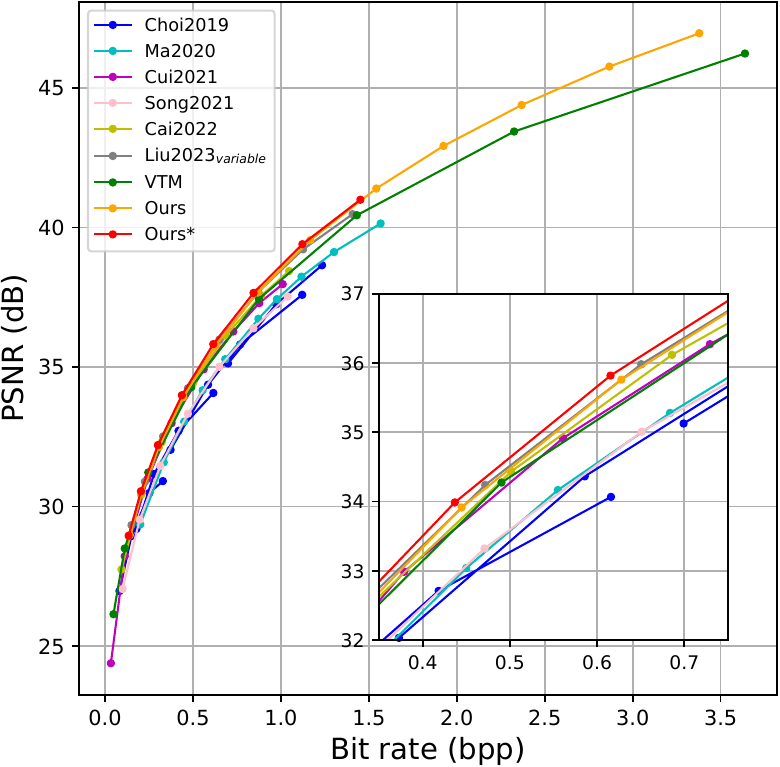}
    \label{fig:kodak_wider}}
    \subfloat[CLIC Pro Valid 2020]{
    \includegraphics[width=0.32\linewidth]{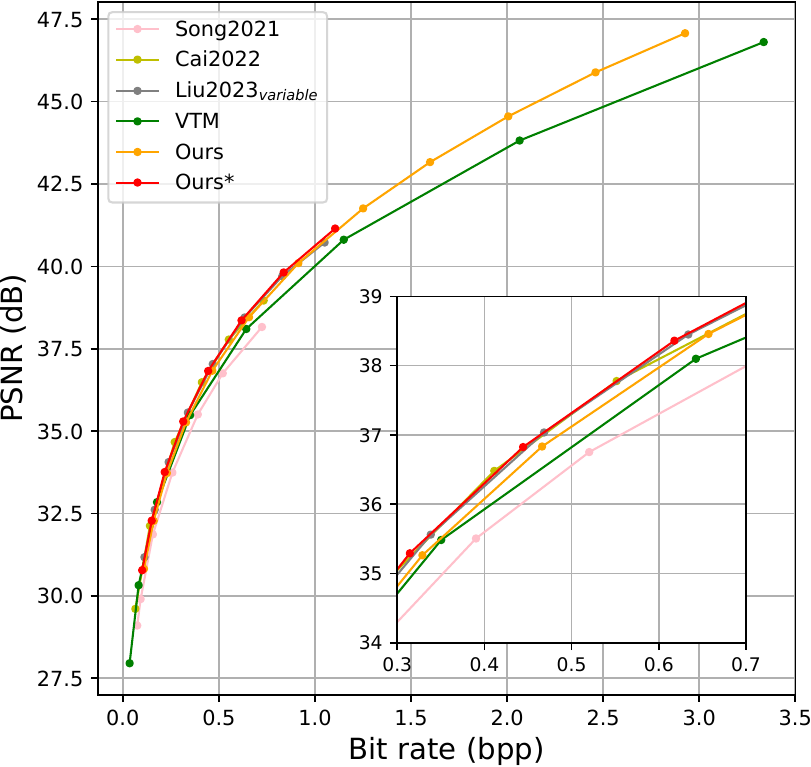}
    \label{fig:clic_wider}}
\subfloat[Tecnick]{
    \includegraphics[width=0.32\linewidth]{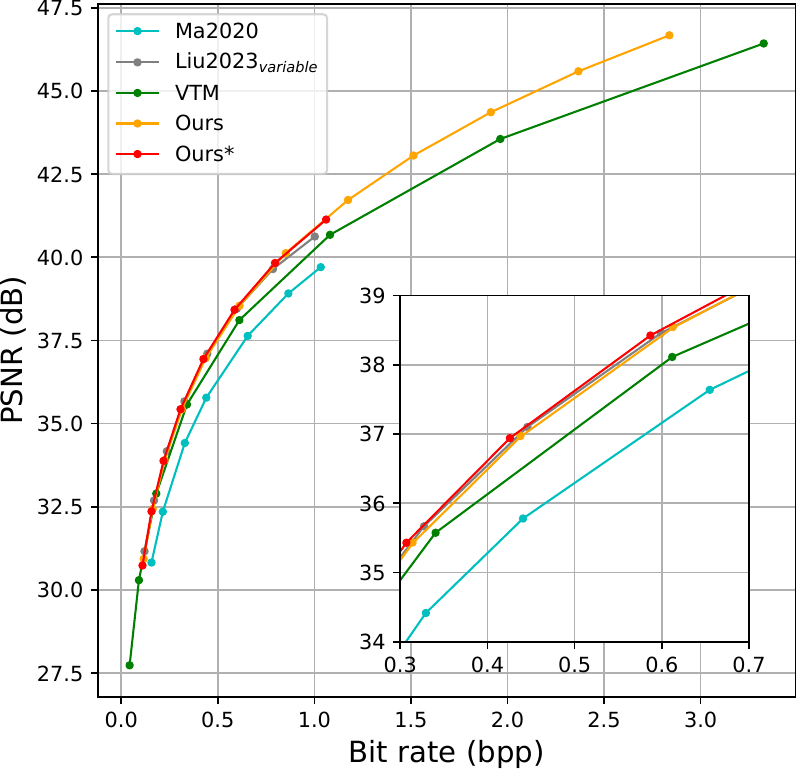}
    \label{fig:tecnick_wider}}
   \caption{Rate-distortion performance comparison between our method and variable-rate image compression methods in a wide range of bit rates. ``Ours'' represents our model with $\lambda_{max}=1.8$, while ``Ours$^*$'' represents our model with $\lambda_{max}=0.18$.}
   \label{fig:rd_wider}
\end{figure*}

 \begin{figure*}[t]
  \centering
   % \includegraphics[width=0.8\linewidth]{fig/mask_conv.pdf}
   \subfloat[Kodak]{
    \includegraphics[width=0.32\linewidth]{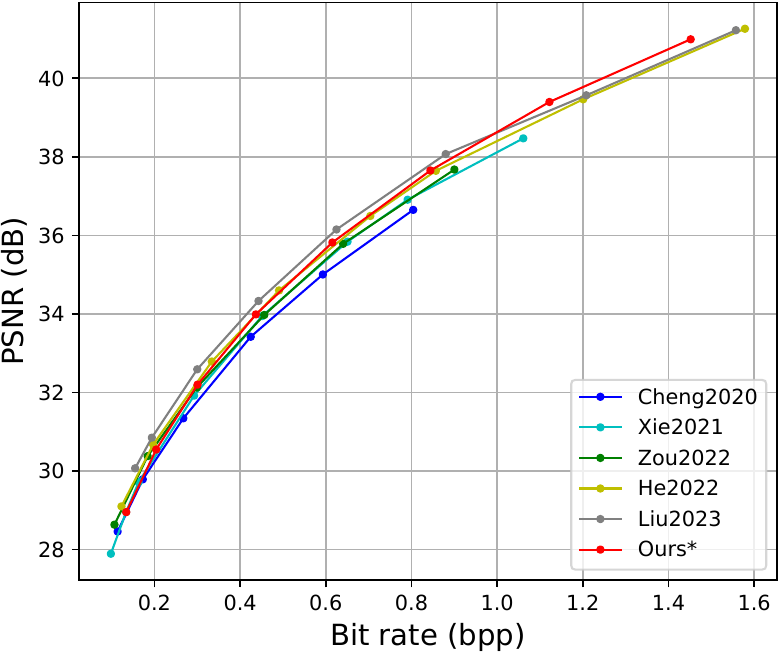}
    \label{fig:kodak}}
     \subfloat[CLIC Pro Valid 2020]{
    \includegraphics[width=0.32\linewidth]{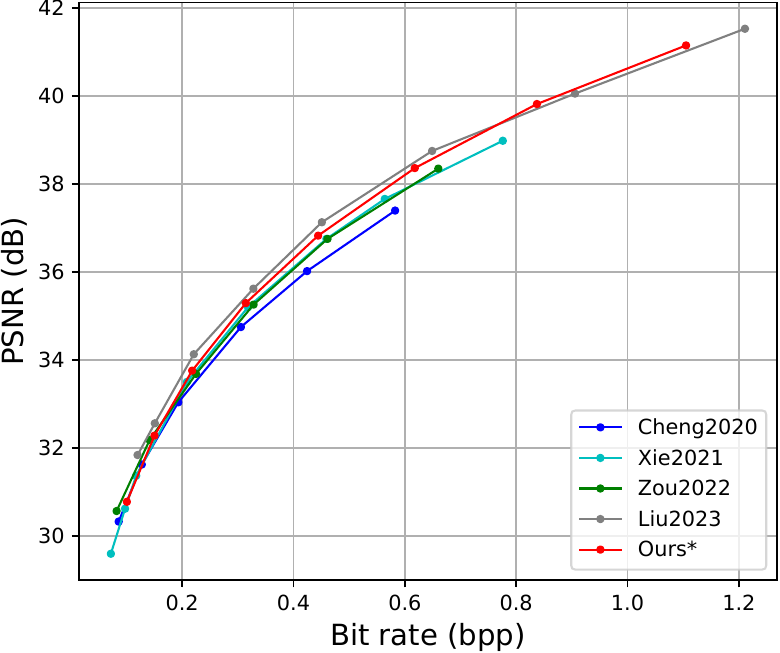}
    \label{fig:clic}}
  \subfloat[Tecnick]{
    \includegraphics[width=0.32\linewidth]{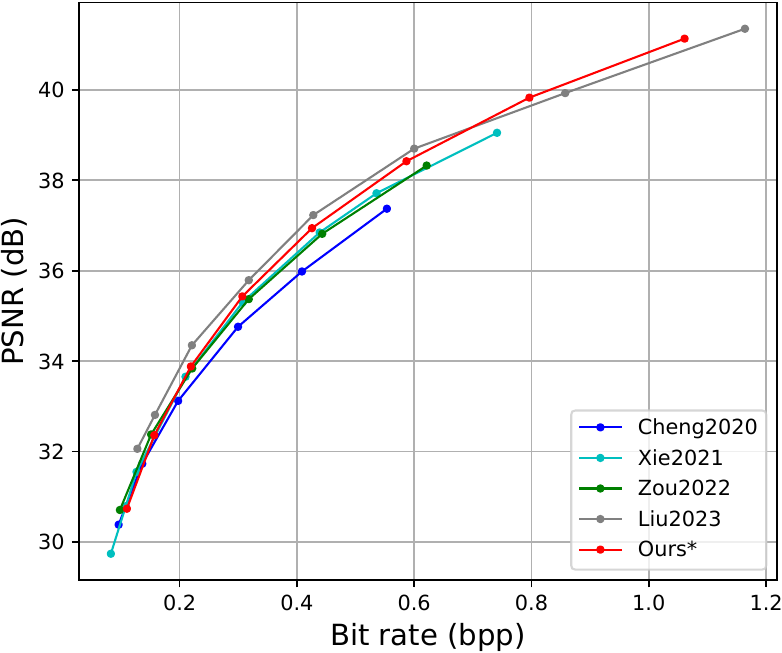}
    \label{fig:tecnick}}
   \caption{Rate-distortion performance comparison between our method and fixed-rate image compression methods. ``Ours$^*$'' represents our model with $\lambda_{max}=0.18$.}
   \label{fig:rd}
\end{figure*}

\section{Experiments}
\label{sec:exp}

\subsection{Experimental Setup} 

\subsubsection{Training Details}

Our variable-rate model is trained on the Flicker2W \cite{liu2020unified} dataset. The training set consists of  20511 images and the remaining 200 images are used for quick validation. During training, We randomly crop $256\times256$ patches and use a batch size of 8. We employ the Adam optimizer \cite{kingma2014adam} with an initial learning rate of $1\times10^{-4}$. The model is trained for a total of 750 epochs (approximately 2M iterations), with the learning rate decreasing to $1\times10^{-5}$ at the 675th epoch and  $1\times10^{-6}$ at the 720th epoch. 

% For ablation study, we train the models for 400 epochs and decay the learning rate at the 300th epoch.

 For the network architecture, we utilize four invertible blocks, each consisting of four invertible units.  For the first three invertible blocks, the channel number of the intermediate layer is set to 128, while for the last invertible block, it is set to 192. The channel number of the intermediate layer in the channel-wise context model, spatial context model, and latent residual prediction model are all set to 128. For the post-processing network, we utilize a lightweight U-Net \cite{ronneberger2015u} containing $1.6M$ parameters.

 For rate adaptation, the max quality value $q_{max}$ is set to 11, and  $\lambda_q$ in \cref{eqn:loss} belongs to $\{$0.0018, 0.0035, 0.0067, 0.0130, 0.0250, 0.0483, 0.0932, 0.1800, 0.320, 0.569, 1.012, 1.8$\}$ for MSE distortion. This model covers a wide range of bit rates similar to the traditional codec VTM. We also train another model with $q_{max}$ equal to 7 ($\lambda_{max}=0.18$) to obtain a comparable bitrate range with recent fixed-rate methods for a fair comparison. Another approach for a fair comparison is training other methods with bigger $q_{max}$. However, training multiple fix-rate models for all other methods is rather time-consuming and we cannot guarantee the optimal training schedule for their methods. We first train the model with a narrower bit rate range and fine-tune it to obtain a wider-range model. The initial learning rate for fine-tuning is set to 1e-5. We find such a fine-tuning process can benefit the RD performance at lower bit rates for the model with a wider bit rate range, compared to training from scratch.

\subsubsection{Evaluation}

 We evaluate our model on three datasets, including Kodak image set \cite{kodak1999} containing 24 images with size of $768\times512$, CLIC professional validation dataset \cite{clic} containing 41 images with $2K$ resolution, and old Tecnick dataset \cite{asuni2014testimages} containing 100 images with size of $1200\times1200$.

 \subsection{Comparison with the SOTA Methods}
 
 \subsubsection{Rate-distortion Performance}
 
 We compare our method with state-of-the-art variable-rate image compression methods \cite{choi2019variable, song2021variable, ma2022end, cui2021asymmetric, cai2022high}, fixed-rate methods \cite{cheng2020learned, xie2021enhanced, he2022elic, zou2022devil, liu2023learned}, and the traditional codec VVC (VTM-12.1) \cite{jvet2020vvc}. Specifically, the methods \cite{ma2022end, xie2021enhanced, cai2022high} are also based on invertible neural networks. For the latest fixed-rate method Liu2023 \cite{liu2023learned}, we modify it to obtain a variable-rate version using the same gain units as in our model and train two additional fixed-rate models at higher rates. The training configuration is the same as ours. The RD performance comparison is shown in \cref{fig:rd_wider} and \cref{fig:rd}.

When compared to variable-rate methods, our model with $\lambda_{max}=1.8$ outperforms the traditional codec VVC across a wide range of rates. In particular, as shown in \cref{fig:rd_wider}, the PSNR gain is more pronounced at higher bit rates. Previous methods only focus on a limited rate range. When only considering low rates, our model with $\lambda_{max}=0.18$ (denoted as Ours$^*$) further enhances the compression performance, surpassing all the existing variable-rate methods.
 
In comparison to fixed-rate methods, our model still shows competitive performance, surpassing most of the existing learned image compression methods, as shown in \cref{fig:rd}. Our model performs slightly worse than the latest method \cite{liu2023learned}, but outperforms it at higher bit rates. This can be attributed to the fact that our model is built on the invertible neural network and the latent representations to be compressed contain all the information of the input image. It is worth noting that we still use our single variable-rate model here.

\begin{figure}[t]
  \centering
   \includegraphics[width=0.7\linewidth]{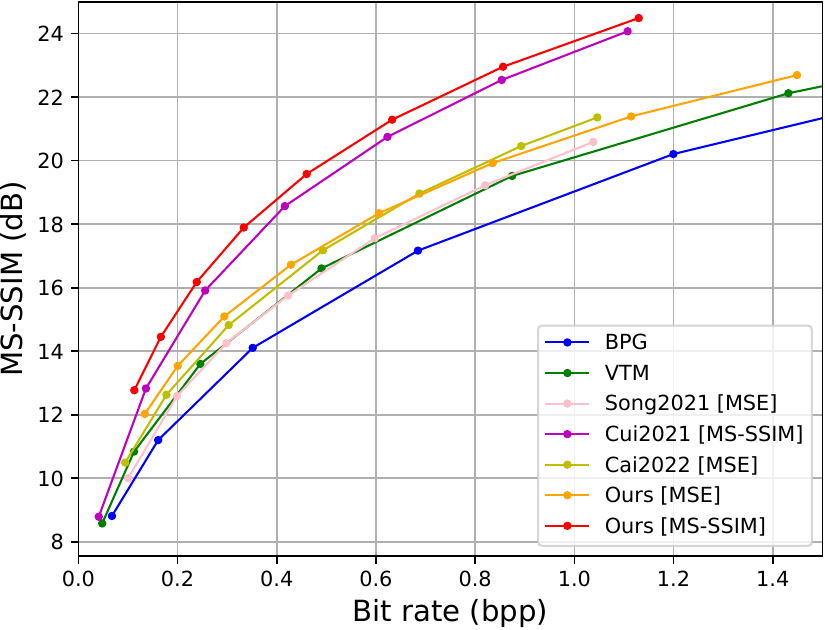}
   \caption{MS-SSIM rate-distortion performance comparison between our method and variable-rate image compression methods on Kodak dataset. MS-SSIM is converted to decibels $-10\log_{10}(1-\text{MS-SSIM})$ for better visual clarity.}
   \label{fig:ssim}
\end{figure}

We also provide the MS-SSIM rate-distortion performance of our model optimized for both MSE and MS-SSIM distortion metrics. The model optimized for MS-SSIM is fine-tuned from the MSE-optimized model with a learning rate of 1e-5. The results on Kodak are shown in \cref{fig:ssim}, where MS-SSIM (dB) is calculated by $-10\log_{10}(1-\text{MS-SSIM}))$. We compare our method with recent variable-rate image compression methods \cite{song2021variable, cui2021asymmetric, cai2022high}, as well as two traditional image codecs VVC (VTM-12.1) and BPG. Some previous methods only report results for MSE-optimized models. Our models optimized for MSE and MS-SSIM both outperform VVC by a large margin, and our model optimized for MS-SSIM achieves the best RD performance among the compared methods.

\begin{figure}[t]
  \centering
   \includegraphics[width=0.7\linewidth]{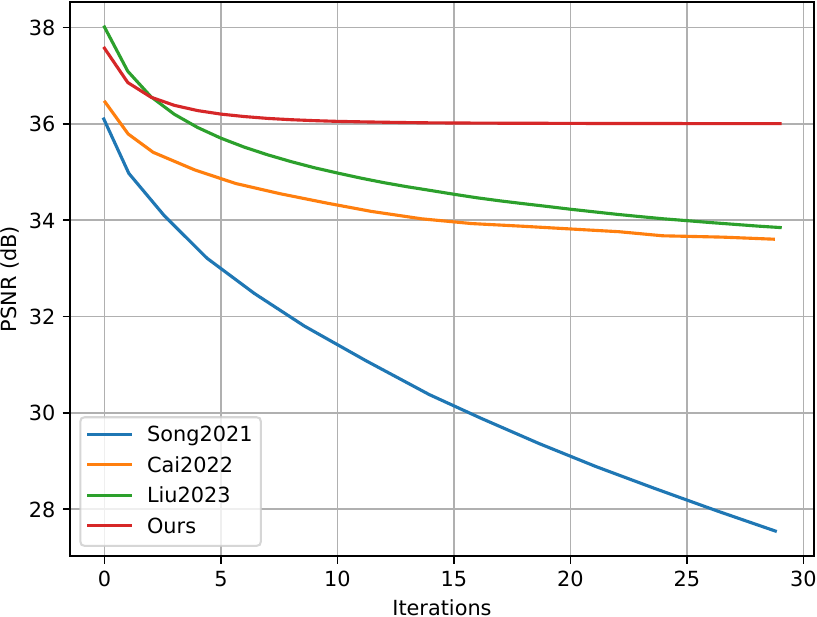}
   \caption{Compression performance after multiple re-encoding operations on Kodak. All models achieve comparable bit rates.}
   \label{fig:reencoding}
\end{figure}

\begin{figure*}[t]
  \centering
   \includegraphics[width=0.9\linewidth]{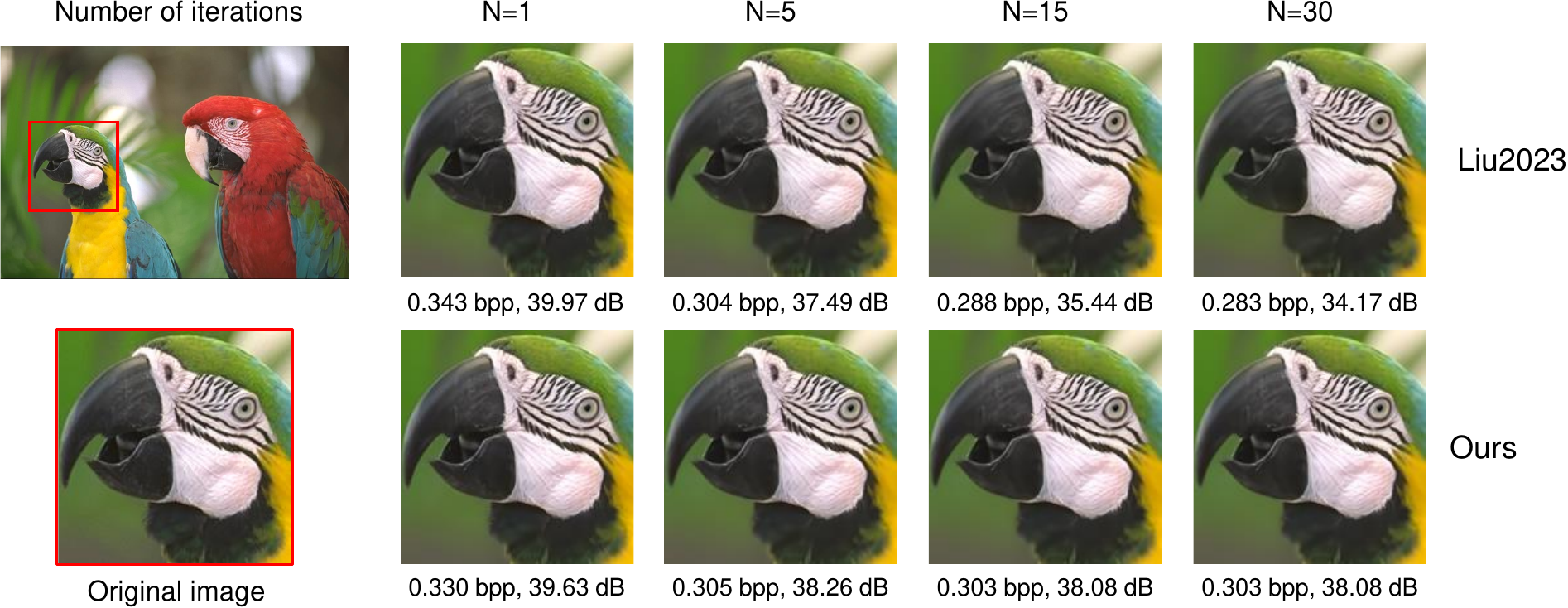}
   \caption{Qualitative results after different re-encoding operations with our method and autoencoder-based method Liu2023 \cite{liu2023learned}.}
   \label{fig:vis_reencoding}
\end{figure*}

\subsubsection{Fidelity of Re-encoding}

Invertible neural network (INN)-based image compression is better at preserving the image quality after multiple re-encoding operations \cite{cai2022high}. To verify the fidelity of re-encoding of our method, We compare it with the state-of-the-art INN-based method \cite{cai2022high} and VAE-based methods \cite{song2021variable, liu2023learned}, as shown in \cref{fig:reencoding}.  Both Song2021 \cite{song2021variable} and Cai2022 \cite{cai2022high} achieve a bit rate of around 0.791 bpp, as reported in \cite{cai2022high}. The initial bit rate of our model is 0.832 bpp, which gradually decreases and stabilizes at 0.786 bpp. The rate of the model from \cite{liu2023learned} starts at 0.869 bpp and gradually decreases to 0.772 bpp. Our method achieves the best performance in maintaining the image quality after multiple re-encoding iterations, which is important in practical applications. We also provide a qualitative comparison between our method and Liu2023 \cite{liu2023learned} in \cref{fig:vis_reencoding}. Our method preserves the texture details even after 30 re-encoding operations. The slight degradation mainly comes from the additional dequantization modules in the decoding process, including LRP and post-processing modules, which have a minor impact on re-encoding performance but are crucial for RD performance.

\subsubsection{Complexity}

We evaluate the complexity and BD-rate against VVC of our method and recent learned image compression methods in \cref{tab:comp}. All the models are tested on Kodak with a single RTX 3090 GPU. 
% Our method has much fewer parameters than the existing methods. 
% Note that we only use a single model for all bit rates while the other methods require multiple models. 
Our method stands out for its significantly fewer parameters and achieves a comparable inference latency against the state-of-the-art methods, \ie, decoding an image with $768\times512$ pixels in 200 ms. In addition, our method excels in terms of training speed. For training, our model processes a batch of 8 images in 35.4 ms, which is approximately one-third the processing time of \cite{liu2023learned}.

The BD rate of our method is higher than \cite{liu2023learned}. However, our model covers a much wider continuous rate range from 0.14 to 3.38 bpp using only a single variable-rate model, while Liu2023 \cite{liu2023learned} requires six separate models and can only handle a narrow rate range from 0.15 to 0.88 bpp. Note that the total size of the model parameters of our method is only $12.34 / (76.57\times6) = 2.6\%$ of that of \cite{liu2023learned}. We attribute this significant advantage in model size to the inherent priors of the invertible neural network.

\begin{table}[t]
  \caption{Complexity and compression performance of learned image compression methods. BD-rate against VVC (VTM-12.1) on Kodak is reported. A lower BD-rate indicates better performance. ``$\times N$'' means that $N$ models are required for different rates.}
  \centering
  % \setlength{\tabcolsep}{3pt}
  % \begin{tabular}{@{}lc@{}}
    \begin{tabular}{@{}lccccc@{}}
    % \begin{tabular}{lcccc}
    \hline
    Method & Enc. (s) & Dec. (s) & Params. (M)$\times N$  & BD-rate (\%) \\
    \hline
    Xie2021\cite{xie2021enhanced} & 2.944 & 7.127 & 50.03$\times$8 & -0.035 \\
    Zou2022\cite{zou2022devil} & 0.146 & 0.151 & 99.86$\times$6 & -3.167  \\
    Liu2023\cite{liu2023learned} & 0.207 & 0.199 & 76.57$\times$6 & -11.481 \\
    Ours & 0.189 & 0.191 & \textbf{12.34}$\times$1 & -5.091 \\
    \hline
  \end{tabular}
  \label{tab:comp}
\end{table}

\begin{table}[t]
  \caption{Ablation studies on Kodak dataset. The anchor is our model with $\lambda_{max}=0.18$}
  \centering
  % \setlength{\tabcolsep}{8pt}
  % \begin{tabular}{@{}lc@{}}
    \begin{tabular}{ccc}
    % \begin{tabular}{lcccc}
    \hline
    Method & Params. (M) & BD-rate (\%) \\
    \hline
    Ours* & 12.339 & 0.00 \\
    \hline
    w/o normalization & 12.338 & +2.23 \\
    w/o invertible $1\times1$ conv & 12.290 & +3.38 \\
    w/ additive coupling & 12.281 & +2.30\\
    \hline
    w/ $5\times5$ masked conv  & 10.654 & +1.57 \\
    w/ $7\times7$ masked conv  & 10.912 & +1.23 \\
    w/ $9\times9$ masked conv  & 11.255 & +1.84 \\
    w/ $11\times11$ masked conv  & 11.684 & +2.09\\
    w/o spatial context & 10.386 & +9.38 \\
    \hline
    w/o post-processing & 10.680 & +5.38 \\
    w/o LRP & 9.703 & +6.96 \\
    w/o post-processing \& LRP & 8.044 & +10.37 \\
    \hline
  \end{tabular}
  \label{tab:ab_study}
\end{table}

\subsection{Ablation Study}

\subsubsection{Settings}

For ablation study experiments, we train each model for 600$K$ iterations. The learning rate starts at $1\times10^{-4}$ and drops to $1\times10^{-5}$ at 500$K$ iterations. All models are evaluated on Kodak dataset. The results are shown in \cref{tab:ab_study}. 

\subsubsection{Analysis of the invertible units} 

Each invertible unit comprises a learnable channel-wise normalization, an invertible $1\times1$ convolutional layer and an affine coupling layer, as shown in \cref{fig:inv_units}. We first remove the normalization layer, leading to performance degradation of 2.23\% BD-rate. The learnable normalization layer stabilizes the training process and improves performance with negligible parameters.

The invertible $1\times1$ convolutional layer in each invertible unit fuses the features channel-wisely.  Previous invertible neural network-based image compression methods \cite{ma2022end, xie2021enhanced} exchanged the order of the two split tensors, \ie, $(\bm{u}_1, \bm{u}_2)\to(\bm{u}_2, \bm{u}_1)$, before coupling layers, which can be seen as reversing the tensor $\bm{u}$ along channel dimension. We replace the invertible $1\times1$ convolutional layer with the reversing operation. The $1\times1$ convolutional layer improves the RD performance by 3.38\% BD-rate, which demonstrates the importance of sufficient information exchange. Note that all the invertible $1\times1$ convolutional layers necessitate $0.049M$ parameters, which account for only 0.4\% of the total parameters.

We use the affine coupling layer to enhance the nonlinearity capacity of the invertible transform. By replacing the affine coupling transform with an additive coupling transform, \ie, removing the scale term in \cref{eqn:inv1} and \cref{eqn:inv2}, the performance degrades by 2.30\% BD-rate. Note that we use a single residual block to generate both the bias and scale parameters, instead of two separate blocks as in \cite{xie2021enhanced}. Therefore, the affine coupling layer is more efficient with similar parameters as the additive coupling layer.

\begin{figure}[t]
  \centering
   \includegraphics[width=0.8\linewidth]{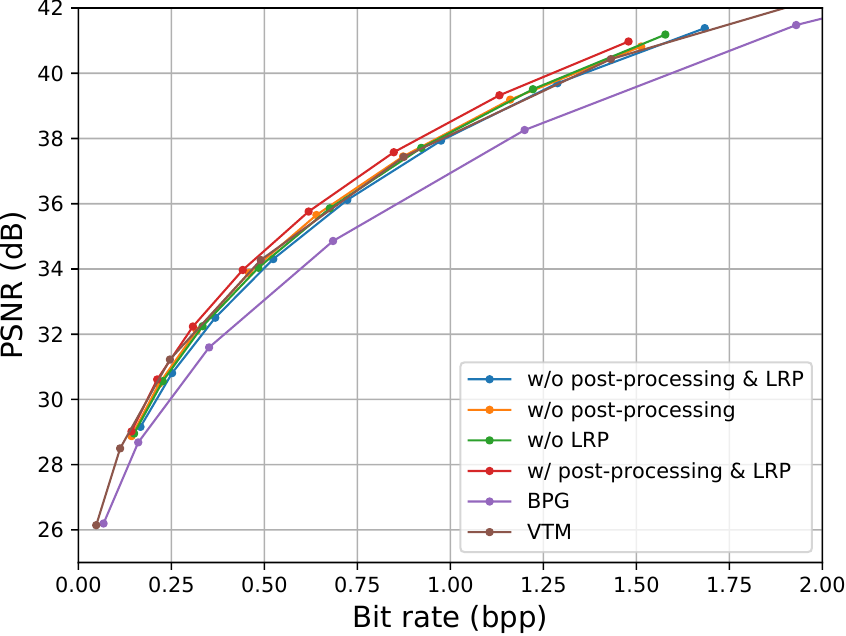}
   \caption{Effects of the post-processing network and the latent residual prediction model.}
   \label{fig:dequan}
\end{figure}

\subsubsection{Analysis of Spatial Context Model} 

To investigate the effect of the proposed multi-layer spatial context model, we develop two variants of our model: one without the spatial context model and another using a single-layer $k\times k$ masked convolutional layer. Removing the spatial context model leads to severe performance degradation, which shows that the spatial correlation is important for accurate entropy modeling of the multi-scale latent representation. A single-layer masked convolutional layer can also enhance the compression performance, but the performance is still not as good as our multi-layer spatial context model. We set the kernel size of the single-layer masked convolutional layer to 5, 7, 9, 11 and observe that the $7\times7$ masked convolutional layer achieves the best performance, which indicates that simply increasing the kernel size of the single-layer convolutional layer can not always improve the RD performance. The stacked 3×3 masked convolutional layers in our proposed multi-layer spatial context model can not only enlarge the receptive field, but also increase nonlinearity while maintaining low complexity. We can adjust the intermediate channel dimensions to enhance the capacity of the context model, which is not achievable with a single convolution layer.

\subsubsection{Analysis of the dequantization modules} 

We evaluate the effects of the dequantization modules, namely the latent residual prediction (LRP) model and the post-processing network by removing them from our model and retraining it. These two modules reduce the quantization error in latent space and pixel space, respectively. It is worth noting that dequantization modules play a crucial role in invertible neural network-based image compression methods. This is because the encoder and the decoder share the same parameters, while in other autoencoder-based image compression methods, the dequantization process is actually integrated within the decoder's parameters. The results are shown in \cref{tab:ab_study} and \cref{fig:dequan}. The RD performance drops significantly without the dequantization modules. Nevertheless, it still achieves comparable performance to VTM and outperforms the BPG by a considerable margin. 
Both the LRP module and post-processing network contribute to enhancing the compression performance. 

\subsubsection{Effect of the number of invertible units} 

\begin{figure}[t]
  \centering
   % \includegraphics[width=0.8\linewidth]{fig/mask_conv.pdf}
   % \subfloat[]{
   %  \includegraphics[width=0.4\linewidth]{fig/inv_conv.pdf}
   %  \label{fig:inv_conv}}
   %   \subfloat[]{
   %  \includegraphics[width=0.4\linewidth]{fig/sp_context.pdf}
   %  \label{fig:sp_context}} \\
  % \subfloat[]{
    \includegraphics[width=0.8\linewidth]{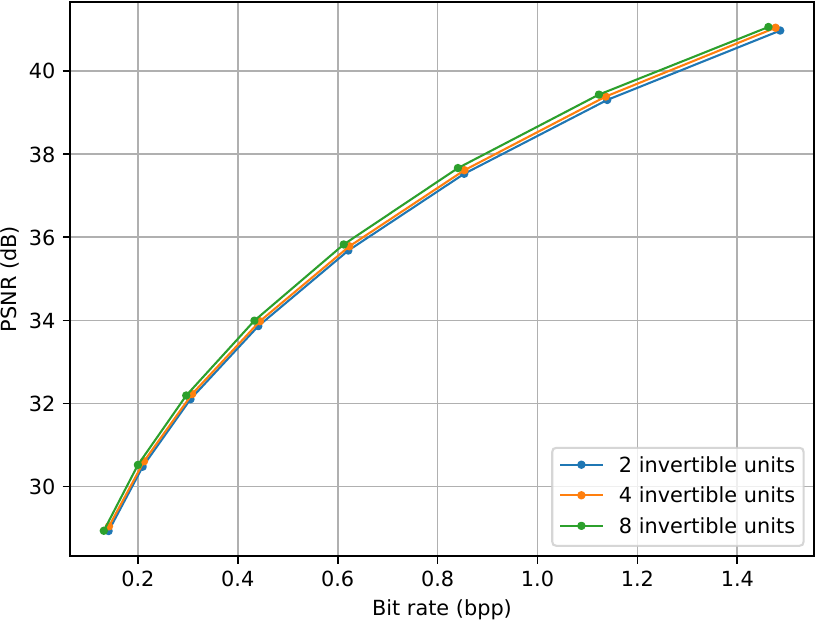}
    \label{fig:inv_depth}
    % }
   \caption{Ablation study on the number of invertible units.}
   \label{fig:ablation}
\end{figure}

The number of invertible units $N$ is set to 4 in each invertible block by default. We also conduct experiments with 2 and 8 invertible units per block to explore the effects of shallower or deeper invertible transforms. The results are shown in \cref{fig:inv_depth}. It can be observed that using 2 invertible units slightly decreases the RD performance. When using 8 invertible units, the RD performance is further enhanced, which suggests that our approach can benefit from more complex transform. Nevertheless, since we aim to develop a lightweight model, we maintain the setting of 4 invertible units in this paper.

\section{Conclusion}
\label{sec:conlusion}

In this paper, we propose a variable-rate image compression method based on a lightweight invertible neural network. Specifically, we design a multi-scale spatial-channel context model to enlarge the receptive field and extend gain units for flexible rate adaptation. Through extensive experiments, we demonstrate that the proposed method outperforms VVC and most existing learned image compression methods, covering a wider rate range with fewer parameters. These results demonstrate the significant potential of employing invertible transform in learned image compression.

% \section*{Acknowledgments}
% This should be a simple paragraph before the References to thank those individuals and institutions who have supported your work on this article.

% {\appendix[Proof of the Zonklar Equations]
% Use $\backslash${\tt{appendix}} if you have a single appendix:
% Do not use $\backslash${\tt{section}} anymore after $\backslash${\tt{appendix}}, only $\backslash${\tt{section*}}.
% If you have multiple appendixes use $\backslash${\tt{appendices}} then use $\backslash${\tt{section}} to start each appendix.
% You must declare a $\backslash${\tt{section}} before using any $\backslash${\tt{subsection}} or using $\backslash${\tt{label}} ($\backslash${\tt{appendices}} by itself
%  starts a section numbered zero.)}


 \bibliographystyle{IEEEtran}
\bibliography{IEEEabrv,ref}

\vfill

\end{document}